%% file: main.tex
\documentclass[10pt,twocolumn,letterpaper]{article}

\usepackage{cvpr}

\usepackage{tcolorbox}
\definecolor{cvprblue}{rgb}{0.21,0.49,0.74}
\usepackage[pagebackref,breaklinks,colorlinks,allcolors=cvprblue]{hyperref}

\def\paperID{*****} 
\def\confName{CVPR}
\def\confYear{2025}

\title{EasyARC: Evaluating Vision Language Models on True Visual Reasoning}

\author{Mert Unsal\\
ETH Zurich\\
{\tt\small mailmertunsal@gmail.com}
\and
Aylin Akkus\\
ETH Zurich\\
{\tt\small aakkus@ethz.ch}
}

\begin{document}
\maketitle
\input{sec/0_abstract}    
\input{sec/1_intro}
\input{sec/3_easy_arc_dataset}
\input{sec/4_evaluations}

\input{sec/5_discussion}
\newpage
{
    \small
    \bibliographystyle{ieeenat_fullname}
    \bibliography{main}
}
\input{sec/appendix}

\end{document}

%% file: sec/0_abstract.tex
\begin{abstract}

Building on recent advances in language-based reasoning models, we explore multimodal reasoning that integrates vision and text. Existing multimodal benchmarks primarily test visual extraction combined with text-based reasoning, lacking true visual reasoning with more complex interactions between vision and language. Inspired by the ARC challenge, we introduce EasyARC, a vision-language benchmark requiring multi-image, multi-step reasoning, and self-correction. EasyARC is procedurally generated, fully verifiable, and scalable, making it ideal for reinforcement learning (RL) pipelines. The generators incorporate progressive difficulty levels, enabling structured evaluation across task types and complexities. We benchmark state-of-the-art vision-language models and analyze their failure modes. We argue that EasyARC sets a new standard for evaluating true reasoning and test-time scaling capabilities in vision-language models. We open-source our benchmark dataset and evaluation code.
\end{abstract}

%% file: sec/1_intro.tex
\section{Introduction}

A long-standing goal of artificial intelligence research
is to develop systems that can reason and answer questions about visual information. Numerous benchmarks assess the visual reasoning capabilities of vision-language models (VLMs). Previous work has created Visual Question Answering (VQA) benchmarks where the model is asked (open-ended) questions about understanding a real world scene generated synthetically or created from real-world data \cite{antol2015vqa} \cite{johnson2017clevr}. Other benchmarks include datasets of geometry problems with image components \cite{lu2021inter} \cite{chen2021geoqa}. However, these benchmarks focus on \textbf{extracting information} rather than \textbf{true visual understanding and reasoning}. In this paper, we introduce \textbf{EasyARC}: a true visual reasoning benchmark inspired from the abstract reasoning challenge ARC \cite{cole2024arc}. We believe EasyARC addresses a crucial gap in multimodal AI: the ability to perform true visual reasoning beyond simple information extraction.

\label{sec:intro}

\begin{figure}[!h]
    \centering
    \includegraphics[scale=0.145]{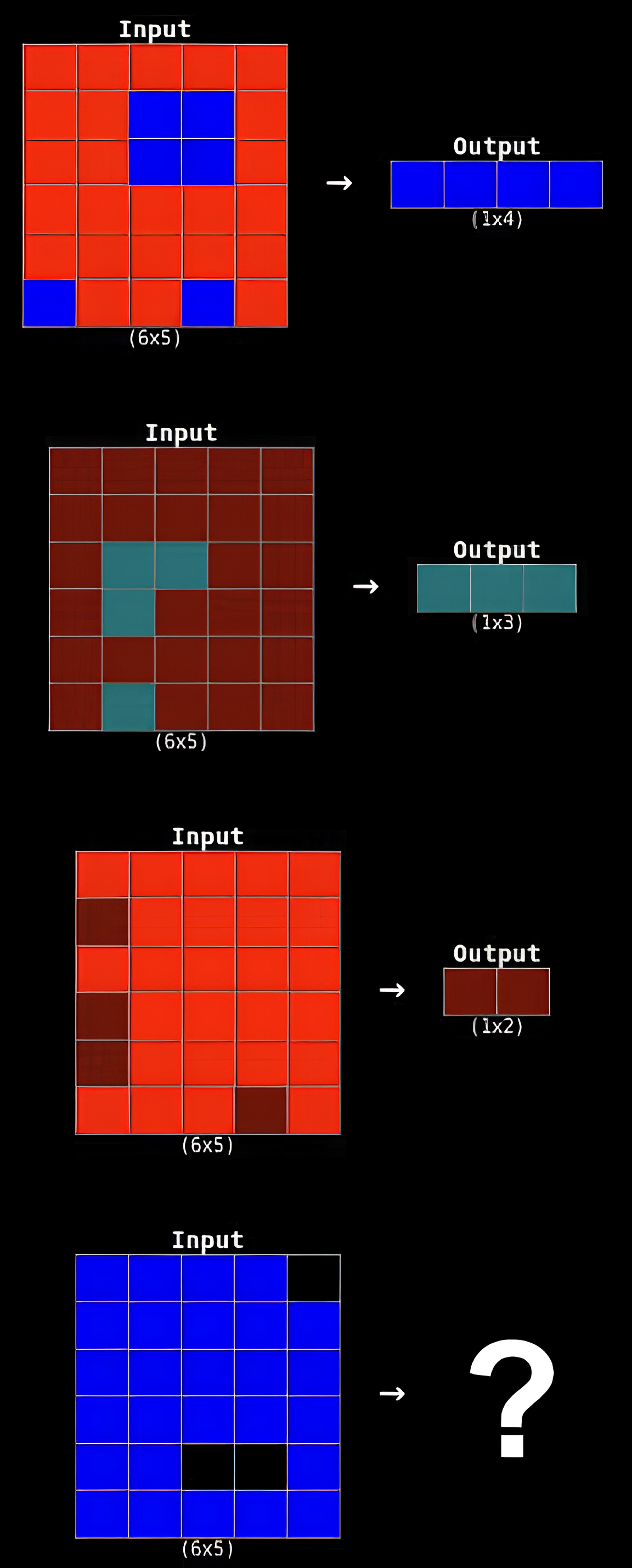}
    \caption{Example Task from EasyARC: The transformation is to identify the largest connected component of non-background color and fill the answer with the component flattened. All SoTA VLMs struggle to understand or solve this example.}
    \label{fig:1}
\end{figure}

The rest of the paper is organized as follows. In \hyperref[sec:2]{Section 2}, we introduce the ARC challenge and consequently EasyARC with our motivation behind creating the dataset. In \hyperref[sec:3]{Section 3}, we evaluate state-of-the-art VLMs on EasyARC, demonstrating that current models fail on this very simple benchmark. Furthermore, we explore \hyperref[sec:failure_modes]{failure modes} of these models in an attempt to understand missing components for this task and plot future research directions. 

%% file: sec/3_easy_arc_dataset.tex
\section{EasyARC Dataset}
\label{sec:2}
\subsection{Original ARC Challenge}
The ARC challenge introduced by \citet{chollet2019measure} is designed as a benchmark for abstract reasoning robust to memorization. It resembles Raven's Progressive Matrices where a matrix of figures are provided and the test taker has to fill in the last missing figure. However, in the \hyperref[fig:1] {ARC setting} we instead have a number of examples consisting of input and output grids which entail a hidden transformation rule between them. In order to demonstrate that the correct transformation rule was found, the test taker then gets one or multiple test input grids and constructs the output grid which is held against a ground truth.

\begin{figure}[h!]
\centering
\includegraphics[width=0.42\textwidth]{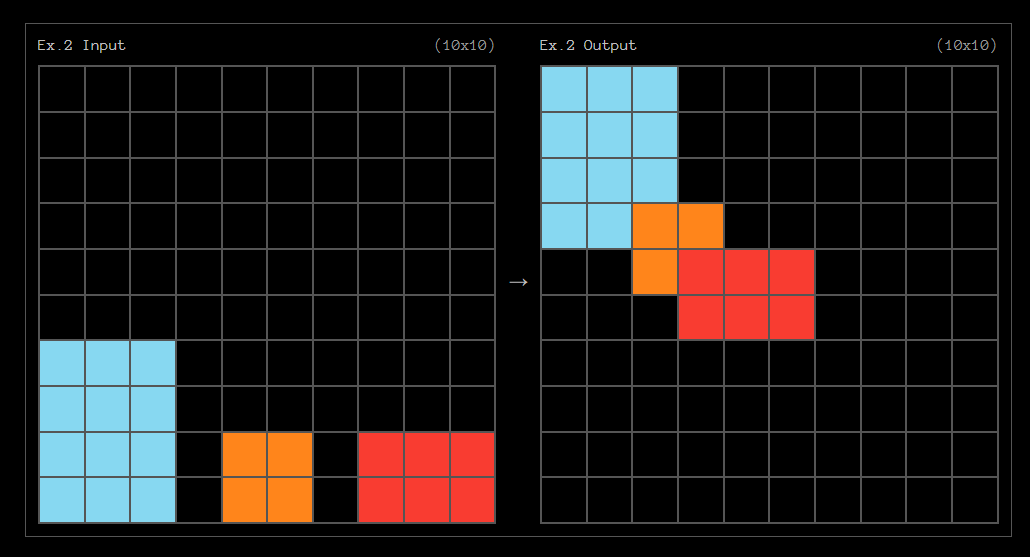}
\caption{Example ARC task from the public evaluation set: Visually, this task is simple as it resembles stacking rectangles in a three-dimensional manner.}
\label{fig:arc_visual}
\end{figure}

ARC originally consists of 400 training and 400 evaluation examples. Common approaches to ARC include (LLM-guided) program synthesis and transductive reasoning with additional test-time-training \cite{li2024combining} \cite{akyurek2411surprising} \cite{li2024programming}. In both cases training examples and test input grids are usually provided directly in the form of language. \citet{li2024combining} demonstrate that these two modes of reasoning are complementary for solving the ARC challenge.
\textbf{Motivated by this, we investigate whether visual reasoning can complement existing approaches to solving ARC problems.} This motivation stems from the observation that humans primarily rely on vision to solve ARC problems. Some patterns become evident in the visual domain but could appear convoluted when described through lengthy natural language representations of the grid.

\subsection{EasyARC}

We initially evaluated VLMs on ARC, expecting them to generalize. However, even state-of-the-art models — both open and closed-source — scored near 0\% when given only images. This failure reveals a fundamental limitation: VLMs struggle with spatial reasoning and grid-based transformations.

In order to facilitate a more meaningful evaluation, we developed EasyARC with tasks that are similar to the original ARC challenge in spirit yet require much simpler spatial/visual understanding. Since ARC challenge has very small training dataset, non-vision approaches to ARC often rely on synthetic data generation \cite{rearc, li2024combining, akyurek2411surprising}. Specifically, \citet{li2024combining} categorized the 400 original training examples into 160 seed tasks, each annotated with human-written Python code for generating examples, a natural language description, and conceptual tags. Taking inspiration from this approach, we designed generators that create tasks in three difficulty levels: easy, medium, and hard. Higher levels typically require a superset of skills from previous levels. Due to the combinatorial possibilities of grid sizes, color permutations, shape positions, axes, and other factors, each individual task generates a multitude of examples.

For initial evaluation, we selected five problem classes that cover a diverse set of concepts:
\begin{itemize}
\item \textbf{Cross, Star:} Introduces the model to basic geometric shapes on grids (e.g., rows, columns, diagonals) and their respective anchor points.
\item \textbf{Counting Cells:} Requires identifying and counting individual cells or contiguous shapes on a uniform background.
\item \textbf{Double Grid:} Involves copying a pattern while applying a horizontal flip or a color flip.
\item \textbf{Dominant Side:} Requires understanding sections of the image (e.g., left or upper side). The transformation colors the entire image in the dominant side's color while filtering out or preserving distracting noise pixels.
\item \textbf{Drop One Color:} Involves transforming one color into another while preserving a third unaffected color. The goal is to determine which color mappings apply.
\end{itemize}

We generate 1000 and 100 random samples for train and test splits from each category, respectively. For all tasks, we sample from the easy, medium, and hard categories with probabilities 0.5, 0.35, and 0.15 and sample number of rows and columns uniformly random between 3 and 10 unless there is a specific requirement for a task. This results in a diverse dataset of 5000 train and 500 test examples. We describe detailed task descriptions in the \hyperref[sec:appendix]{Appendix} with images included.

\subsection{Benchmark for Test-Time Scaling of VLMs}

EasyARC serves as an ideal test-bed for test-time scaling in vision models. In particular, EasyARC is:

\begin{enumerate}
\item \textbf{True Reasoning Benchmark:} EasyARC tasks require multi-step reasoning, including hypothesis formation (e.g., "Predict minority cell color"), iterative validation (e.g., "But example 2 has a 1x2 blue grid instead of a single color"), hypothesis refinement, and alternative perspective-taking.
\item \textbf{Synthetic and Easily Expandable:} EasyARC allows flexible dataset creation via configurable parameters such as grid sizes and difficulty probabilities. Our dataset includes 5000 training and 500 evaluation tasks, which we will release alongside the generators. Additionally new generators can be seamlessly integrated.
\item \textbf{Verifiable:} Since the task requires an exact output grid prediction, evaluation is straightforward. This avoids reliance on proxies such as open-ended VQA, where complex multi-step reasoning tasks often sacrifice verifiability.
\item \textbf{Gradual Difficulty Levels:} We can adjust task difficulty progressively to provide a structured learning signal in an RL pipeline, allowing for insights into training dynamics.
\item \textbf{Multi-Image Reasoning:} Each example consists of multiple input-output image pairs. The model must correctly associate pairs, attend to their transformations, describe them, and generalize accordingly.
\end{enumerate}

\textbf{These characteristics make EasyARC particularly well-suited for reinforcement learning (RL) pipelines where the base problem dataset's difficulty distribution is of utmost importance. }

%% file: sec/4_evaluations.tex
\section{Evaluation of SoTA VLMs}
\label{sec:3}
In this section, we present our findings on the performance of SoTA VLMs on our EasyARC benchmark. We evaluate closed-source models such as Claude 3.7 Sonnet and GPT4o-mini as well as small and large open source models such as QVQ-72B-Preview, Qwen2.5-VL-7B-Instruct, and Gemma3-27B-Instruct. We pass a stacked image of input output grids as well as a text prompt describing the task to all models. We use a low temperature of 0.5 and collect 1 sample from each model due to budget constraints. 
Despite the fact that our benchmark is designed to be extremely easy, we find that all models except Claude 3.7 Sonnet exhibit weak performance on this task, achieving scores below \%20. We share the prompt in the \hyperref[sec:appendix]{Appendix}.

\begin{figure}[!h]
    \centering
    \includegraphics[width=1.1\linewidth]{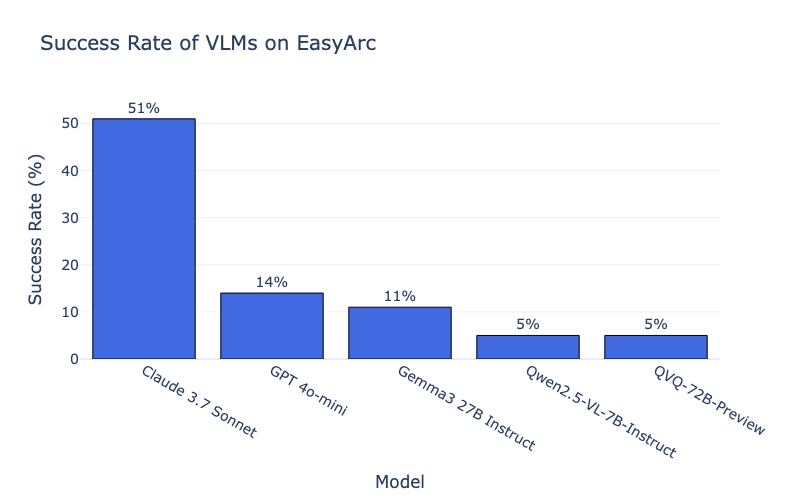}
    \caption{Success Rate of VLMs on EasyARC}
    \label{fig:enter-label}
\end{figure}

We could not evaluate OpenAI o-series models extensively since the API access doesn't allow images as of the time we are writing this paper. Through our manual evaluation, we expect o3-mini to score somewhere between Claude 3.7 Sonnet and GPT4o-mini. 

\subsection{Failure Modes}
\label{sec:failure_modes}
To better understand the challenges faced by VLMs on EasyARC, we manually analyze their outputs. 

\subsection{What Can Claude See?}

Claude 3.7 Sonnet is the only model that demonstrates a degree of true visual reasoning. To analyze its limitations, we break down its success rate by problem category.

\begin{figure}[h!]
    \centering
    \includegraphics[width=\linewidth]{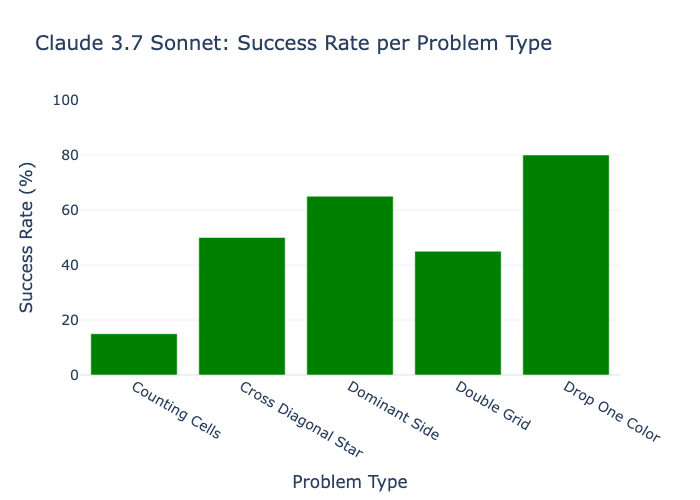}
    \caption{Claude 3.7 success rate across problem types.}
    \label{fig:claude}
\end{figure}

To interpret this plot meaningfully, recall that each category contains problems sampled with probabilities 0.5, 0.35, and 0.15 from the easy, medium, and hard levels, respectively. We observe that the model struggles the most with Counting Cells, which requires precisely counting connected components of a specific color. In contrast, it performs significantly better on Drop One Color, where it simply identifies a missing color, and Dominant Side, where it determines which side of the image dictates the output color.

These results suggest that the model perceives a “blurry” representation of images—capturing high-level features while missing fine-grained details. Notably, most of its successes come from easy tasks; as soon as noise is introduced or the mapping becomes more complex, its performance deteriorates. This highlights a fundamental limitation in its ability to generalize beyond simple pattern recognition.

\subsection{Task Analysis: Counting Cells}

Below, we include an example of an example input output for the Counting Cells task, where the model needs to simply find the largest connected component for the minority color and count the number of cells in this component. All models, including the most successful Claude 3.7 Sonnet, failed at understanding this transformation. 

\begin{figure}[h!]
    \centering
    \includegraphics[width=0.5\linewidth]{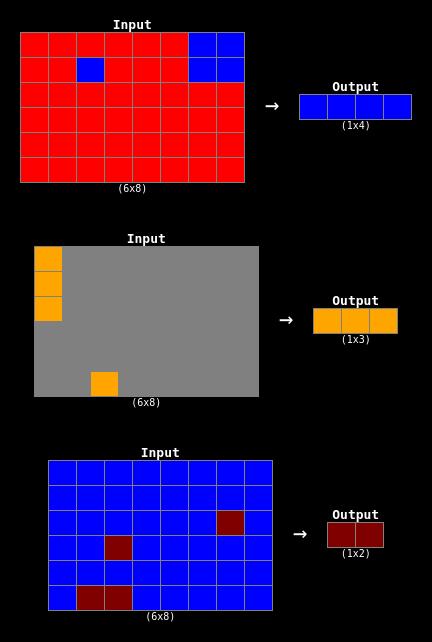}
    \caption{Example Input-Output for Counting Cells Task}
    \label{fig:enter-label}
\end{figure}

The model usually identifies that the output should be of minority color, which allows it to solve the easy version of this task. However, the model is unable to look at the image "longer" to realize that it should find and count the largest connected component. \textbf{This failure suggests that test-time compute on images may be necessary, as models rely solely on the training examples to infer transformation rules rather than processing the image step-by-step.}
\subsection{Task Analysis: Cross/Star}
While many tested models successfully solved the Cross/Star task, we identified two notable failure modes.

The first failure mode is related to the model’s inability to precisely locate objects such as cells, rows, columns, and diagonals. Specifically, the model often fails to recognize that the highlighted cell serves as the anchor point for the transformation. This limitation appears to stem from two potential bottlenecks: On the one hand, the vision component may struggle to establish a direct correspondence between the cell's position in the input and the placement of the resulting row in the output. On the other hand, if the model was capable of extracting precise positional information per grid in natural language (instead of comparing them directly visually), it could potentially mitigate this issue; however, our findings suggest that it lacks this capability as well.

The second failure mode concerns the generation of diagonal structures. Even when the model correctly identifies the transformation rule and the orientation of the diagonal (e.g., upper-right direction), it frequently fails to fully extend the diagonal. In many cases, the model only completes the upper half of the diagonal after the anchor point while neglecting the preceding portion. Additionally, some models generate the diagonal in an incorrect position, despite correctly articulating that it should be anchored at the given input cell. These findings suggest that the model struggles with effectively integrating information across modalities—specifically, transferring spatial knowledge from the visual input to linguistic reasoning and subsequently applying it correctly in the output grid.
\subsection{Analysis of QVQ-72B-Preview Reasoning Model}

QVQ-72B-Preview is the only reasoning model that was trained with reinforcement learning to solve problems in our benchmark, in a similar fashion to o-series models by OpenAI. When we look at the outputs of the model, we realize that model has a clear mode of reasoning: even though the model is supposed to do visual reasoning, it follows:

\begin{enumerate}
    \item Extract the input output images into text.
    \item Do text based reasoning to solve the problem.
\end{enumerate}

Although correctly extracting the images is necessary to solve the task, we also expect the model to mix the vision and language understanding rather than using the image only for extraction purposes. We speculate that this is an artifact of the model being trained on datasets such as Geometry3k \cite{lu2021inter} where it is sufficient to extract and solve the problem to get the reward. On the contrary, EasyARC presents a benchmark where the VLM has to interact with the image and do some true visual reasoning rather than mere extraction.

%% file: sec/5_discussion.tex
\section{Conclusion}

We introduced EasyARC, the first benchmark designed to evaluate true visual reasoning in vision-language models. Unlike existing benchmarks that focus on visual extraction, EasyARC requires multi-step reasoning, hypothesis formation, and iterative validation. Our results show that state-of-the-art VLMs struggle with even simple EasyARC tasks, highlighting fundamental gaps in visual cognition. With its scalable data generation and structured difficulty levels, EasyARC provides a valuable testbed for studying test-time scaling and reinforcement learning strategies in visual reasoning. We release our dataset and evaluation code to foster further research in this direction.

%% file: sec/appendix.tex
\clearpage
\setcounter{page}{1}
\maketitlesupplementary

\section{Appendix A: Details of the tasks}

\label{sec:appendix}
\subsection{Cross/Star}

The \textbf{Cross/Star} task involves generating structured patterns based on highlighted cells. The complexity increases as follows:

\begin{itemize}
    \item \textbf{Easy:} Given a uniform background with a single distinct cell, the model must generate a row, column, or diagonal anchored at that cell.
    \item \textbf{Medium:} The model must create a more complex shape, such as a cross (combining a row and column) or both diagonals intersecting at the given cell.
    \item \textbf{Hard:} Two distinct cells, each in different colors, are provided. The model must replicate the medium-level shapes while following a more intricate coloring scheme.
\end{itemize}

\textbf{Key Concepts:} Geometric structures (rows, columns, diagonals), pattern formation, and color-based rules.

\subsection{Counting Cells}

This task focuses on counting and identifying spatial properties of highlighted regions.

\begin{itemize}
    \item \textbf{Easy:} Given a uniform background and \( n \) distinct-colored cells, the model must count these cells and output a \( 1 \times n \) grid of the same color.
    \item \textbf{Medium:} The grid contains clusters of connected components, and the model must output the number of cells in the largest such cluster.
    \item \textbf{Hard:} The background becomes striped, making it harder to distinguish and count the target clusters.
\end{itemize}

\textbf{Key Concepts:} Object counting, recognizing clusters, handling background distractions.

\subsection{Double Grid}

This task requires duplicating a grid while introducing transformations.

\begin{itemize}
    \item \textbf{Easy:} Given a grid with a background color and a pattern (dot or stripe), the model must duplicate the grid either horizontally or vertically.
    \item \textbf{Medium:} The duplication is accompanied by a transformation, such as a 180-degree rotation or a color flip.
    \item \textbf{Hard:} A randomly generated \( 3 \times 3 \) grid must be replicated a variable number of times both horizontally and vertically, with transformations such as 90-degree rotations or background color flips.
\end{itemize}

\textbf{Key Concepts:} Duplication, transformation, rotation, color manipulation.

\subsection{Dominant Side}

This task involves detecting and analyzing the dominant portion of a grid.

\begin{itemize}
    \item \textbf{Easy:} A grid is divided into two sides (e.g., left/right, top/bottom). One side (across examples) dictates the output grid's color.
    \item \textbf{Medium:} A uniform noise is introduced, requiring the model to distinguish the dominant side despite distractions.
    \item \textbf{Hard:} The noise becomes more complex, containing multiple colors, while the model must still correctly identify the dominant region.
\end{itemize}

\textbf{Key Concepts:} Spatial division, handling noise.

\subsection{Drop One Color}

This task focuses on recognizing color transformation rules.

\begin{itemize}
    \item \textbf{Easy:} Three colors are chosen, with one always transforming into another while remaining consistent across examples.
    \item \textbf{Medium:} Two colors are fixed, with one transforming into the other, while a third color varies across examples.
    \item \textbf{Hard:} One color transforms into the background, with the background itself varying but making up at least 70\% of the grid.
\end{itemize}

\textbf{Key Concepts:} Color mapping, background detection, transformation rules.

\section{Appendix B: Evaluation}
\subsection{Prompt}
We use the following prompt for evaluating VLMs.

\begin{tcolorbox}[colframe=black,colback=white,boxrule=0.5mm,arc=2mm]
Here is an image of multiple input-output grids. The cells of the grids can be of color black, blue, red, green, yellow, grey, pink, orange, teal, maroon. Your goal is to find the common rule that maps the input grid to the output grid. Make sure that you find an unambiguous transformation rule.

Below is a test input grid. After you find the transformation rule, apply it rigorously step by step to the test input grid to find the output grid. Put the final grid in \texttt{```} in the same format as the input grid, where rows are separated by newline and columns are separated by column.

\begin{verbatim}
{test_input}
\end{verbatim}
\end{tcolorbox}
The image contains a variable number of input-output demonstrations. To ensure a unique solution, some tasks include more examples than others, which can be flexibly adjusted in the generator functions. To guarantee sufficient resolution for recognizing relevant details, the image resolution was set appropriately. The model was also able to read small annotations such as grid shapes and labels like "Input" and "Output", indicating that it had access to a detailed visual representation sufficient for solving the task.
\begin{figure}
    \centering
    \includegraphics[width=0.4\linewidth]{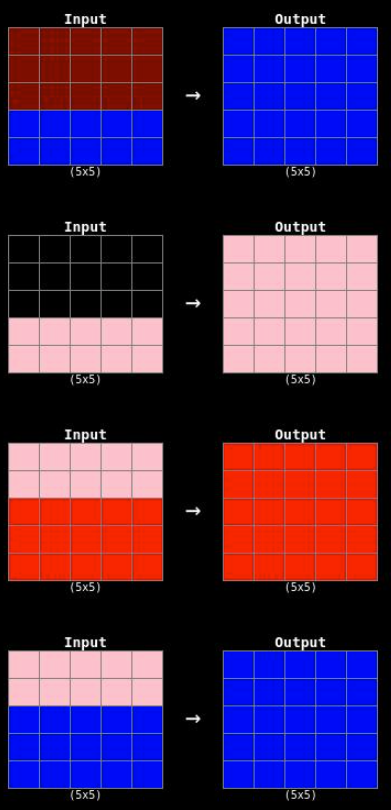}
    \caption{The goal is to predict the color of the lower side of the image, not the majority or minority color. To guarantee uniqueness of the solution we make sure to include diverse demonstrations.}
    \label{fig:enter-label}
\end{figure}
\section{Example Reasoning traces}
In the following section we give an example for a successful and unsuccessful reasoning trace to give some insight into the 
\subsection{Successful Reasoning trace}
This reasoning trace belongs to a problem of the category counting cells with difficulty easy. The model evaluated is Claude 3.7 and it correctly understood that it has to count the number of minority color cells. A common failure mode of other models was that they often only predicted the minority color instead of counting, despite the demonstrations clearly showing this rule:\\
\begin{tcolorbox}[colback=white, colframe=black,
  sharp corners, boxrule=0.5pt, fontupper=\ttfamily,
  width=\columnwidth]
I'll analyze the transformation rule from the input-output examples and then apply it to the test grid.\\

Looking at the examples in the image:\\

1. First example: A red grid (6x6) with one yellow cell produces a yellow output of size 1x2.\\
2. Second example: A green grid (6x6) with one teal cell produces a teal output of size 1x1.\\
3. Third example: A green grid (6x6) with two gray cells produces a gray output of size 1x2.\\
The transformation rule appears to be:\\
- The output grid has the same color as the non-dominant color in the input grid\\
- The size of the output depends on the number of non-dominant colored cells in the input\\
- If there is 1 different cell, the output is 1x1\\
- If there are 2 different cells, the output is 1x2\\

Now applying this rule to the test input grid:\\
- The dominant color is orange\\
- There is one non-dominant cell which is maroon (at position 3,3)\\
- Since there is 1 different cell, the output should be a 1x1 grid with maroon color\\
```\\
maroon\\
```\\
\end{tcolorbox}
\subsection{Unsuccessful Reasoning trace}
This example shows an unsucessful attempt of Claude 3.7 to solve a problem of the category Cross/Star task with difficulty easy. 
\begin{tcolorbox}[colback=white, colframe=black,
  sharp corners, boxrule=0.5pt, fontupper=\ttfamily,
  width=\columnwidth]
    "Looking at the examples in the image, I can see a clear transformation pattern. In each case, there's a single colored cell (pink or red) in the input grid that's different from the background. In the output grid, this cell creates a diagonal line from its position to the bottom-right corner of the grid.
    The transformation rule is:\\
    1. Identify the position of the unique colored cell in the input grid\\
    2. Create a diagonal line of that same color starting from the unique cell and extending to the bottom-right corner\\
    3. The diagonal moves one cell right and one cell down with each step\\
    Let me apply this rule to the test input grid:\\
    1. The unique colored cell is yellow at position (7,4) (row 8, column 5)\\
    2. I need to create a diagonal line of yellow cells from this position to the bottom-right corner\\
    Starting from the yellow cell at (7,4), I'll create a diagonal line moving down-right:\\
    - Position (7,4): yellow (already exists)\\
    - Position (8,5): should be yellow
    The final grid would be:\\
    ```\\
    green green green green green green\\
    green green green green green green\\
    green green green green green green\\
    green green green green green green\\
    green green green green green green\\
    green green green green green green\\
    green green green green green green\\
    green green green green yellow green\\
    green green green green green yellow\\
    ```\\
\end{tcolorbox}
This example illustrates a common failure mode in the model's visual processing: the model identifies the diagonal in the output as extending only in a single direction (toward the bottom right), whereas it is in fact a full diagonal originating from the anchor point.

\subsection{Human Baseline}
Another natural comparison beside the model to model comparison given in figure 3 is the human baseline performance. Human labeling on our dataset has not yet been employed, however human labeling has been performed on the much more difficult ARC dataset which inspires our problem categories and average performance was found to be between 73.3\% and 77.2\% \cite{legris2024h}. We therefore expect that the human baseline is probably significantly above that percentage.
\clearpage
\begin{figure*}[t]
    \centering
    \begin{subfigure}{0.32\textwidth}
        \includegraphics[width=\textwidth]{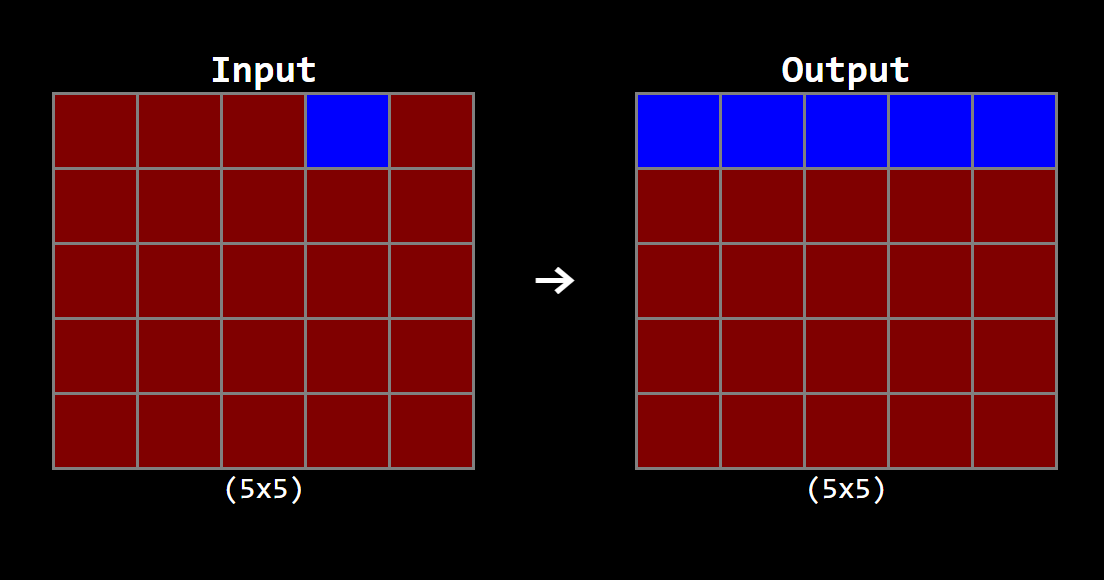}
        \caption{Basic row placement from a single anchor point.}
        \label{fig:easy_cross}
    \end{subfigure}
    \hfill
    \begin{subfigure}{0.32\textwidth}
        \includegraphics[width=\textwidth]{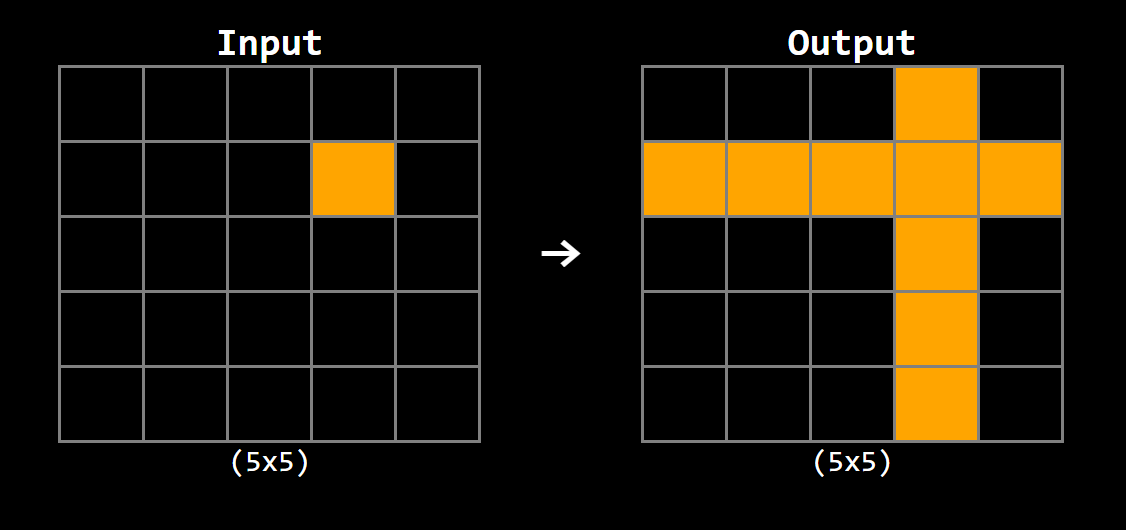}
        \caption{Creating cross at anchor point.}
        \label{fig:medium_cross}
    \end{subfigure}
    \hfill
    \begin{subfigure}{0.32\textwidth}
        \includegraphics[width=\textwidth]{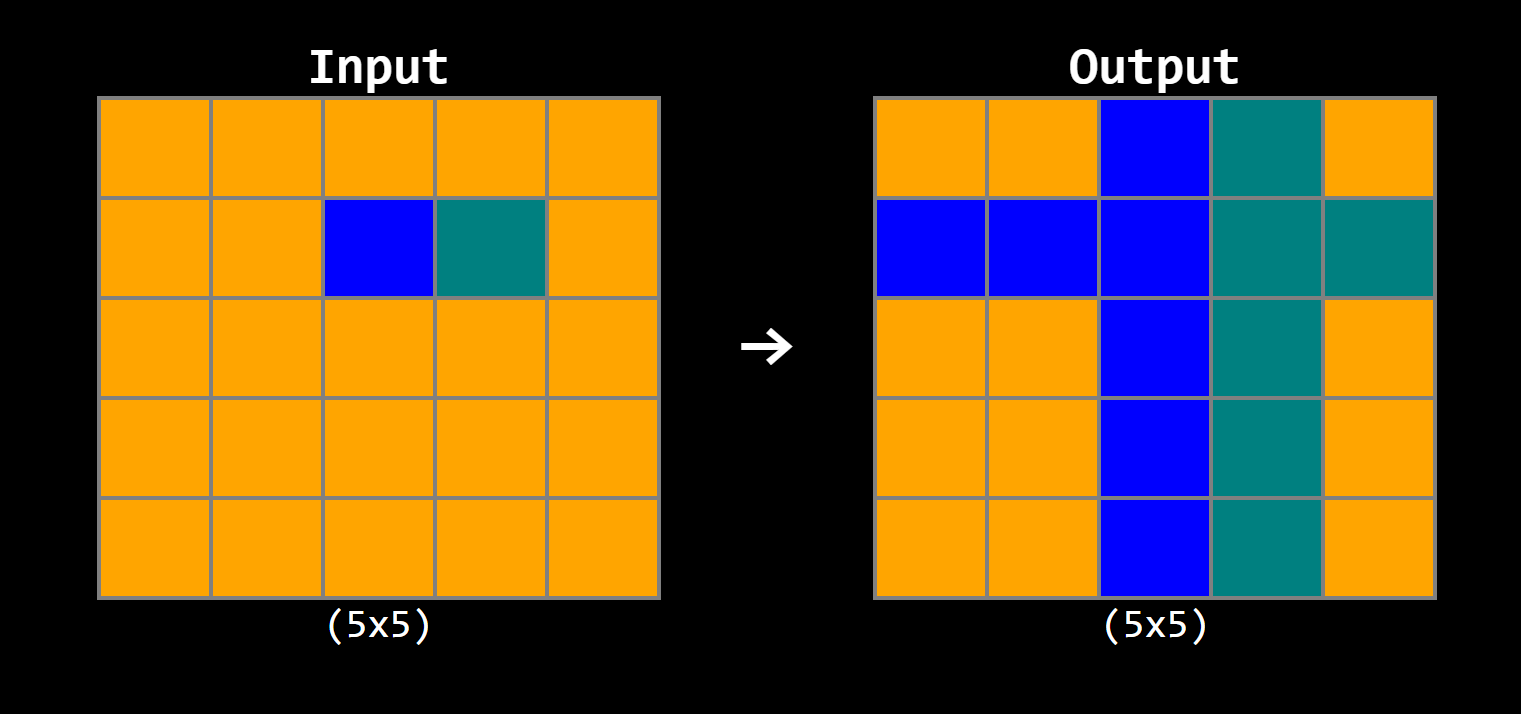}
        \caption{Multiple anchors with more complex color constraints.}
        \label{fig:hard_cross}
    \end{subfigure}
    \caption{Cross/Star Task: The complexity increases as more lines and colors are introduced.}
    \label{fig:cross_comparison}
\end{figure*}

\begin{figure*}[t]
    \centering
    \begin{subfigure}{0.32\textwidth}
        \includegraphics[width=\textwidth]{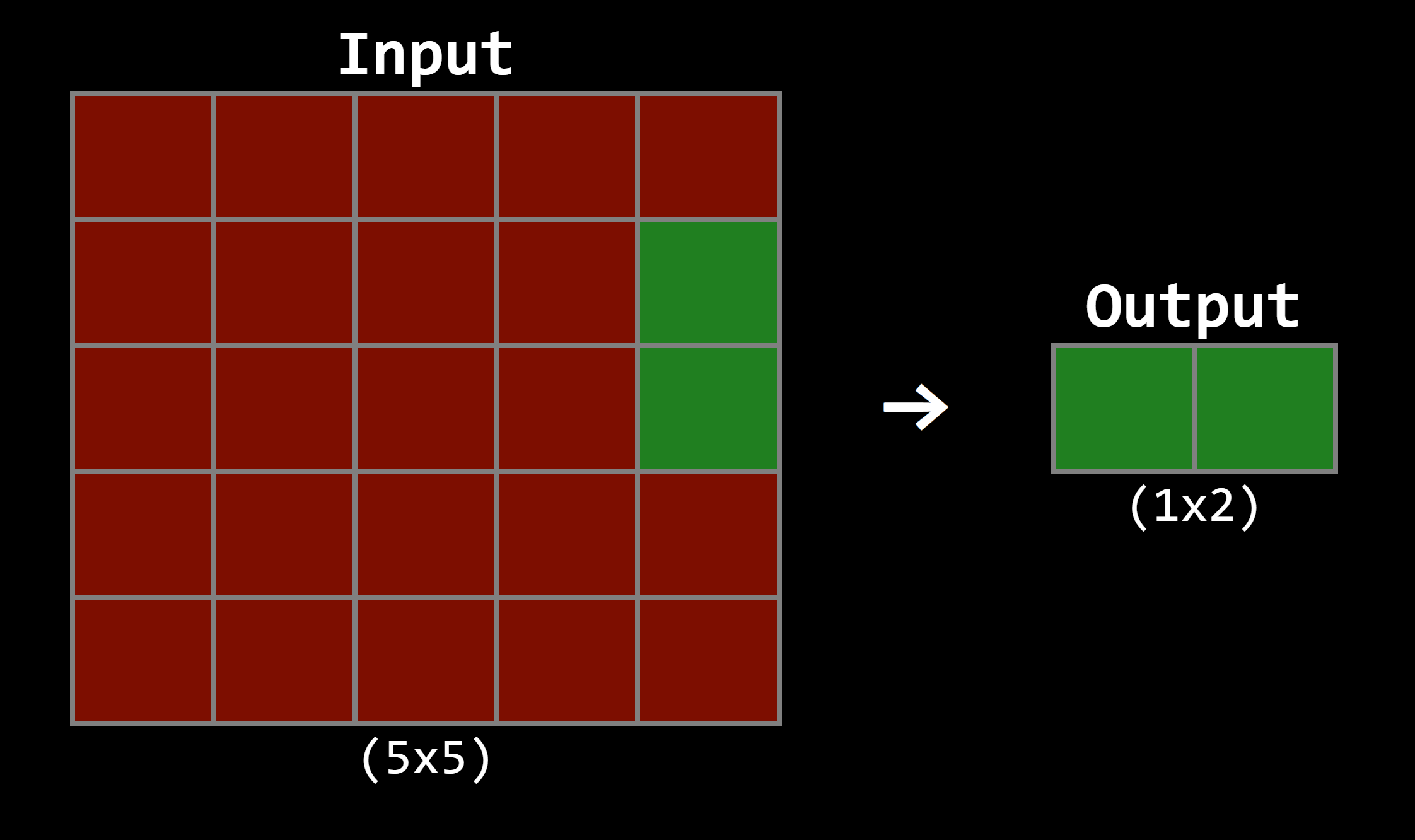}
        \caption{Basic counting of all differently colored cells.}
        \label{fig:easy_count}
    \end{subfigure}
    \hfill
    \begin{subfigure}{0.32\textwidth}
        \includegraphics[width=\textwidth]{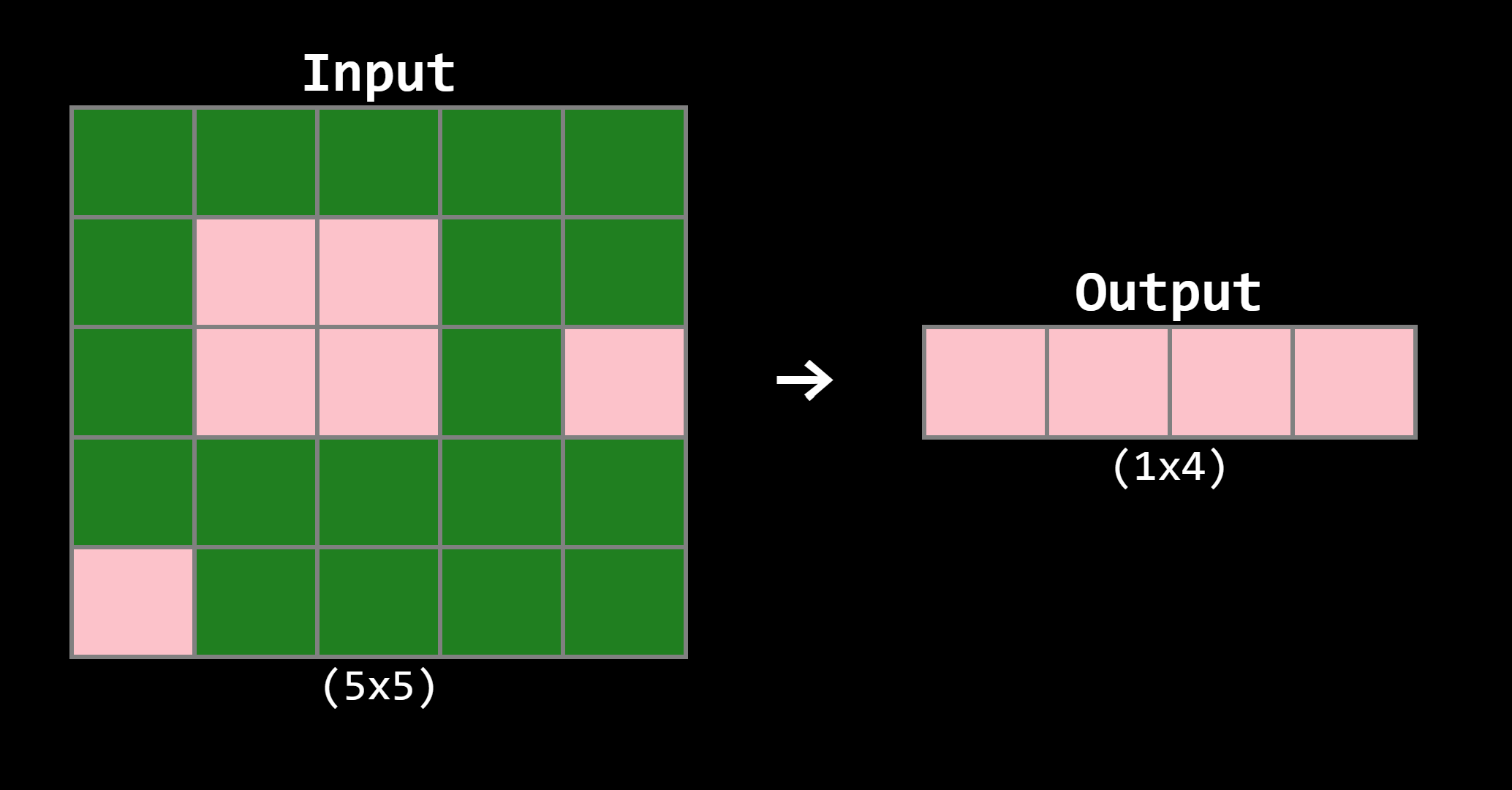}
        \caption{Detecting and counting the largest connected component.}
        \label{fig:medium_count}
    \end{subfigure}
    \hfill
    \begin{subfigure}{0.32\textwidth}
        \includegraphics[width=\textwidth]{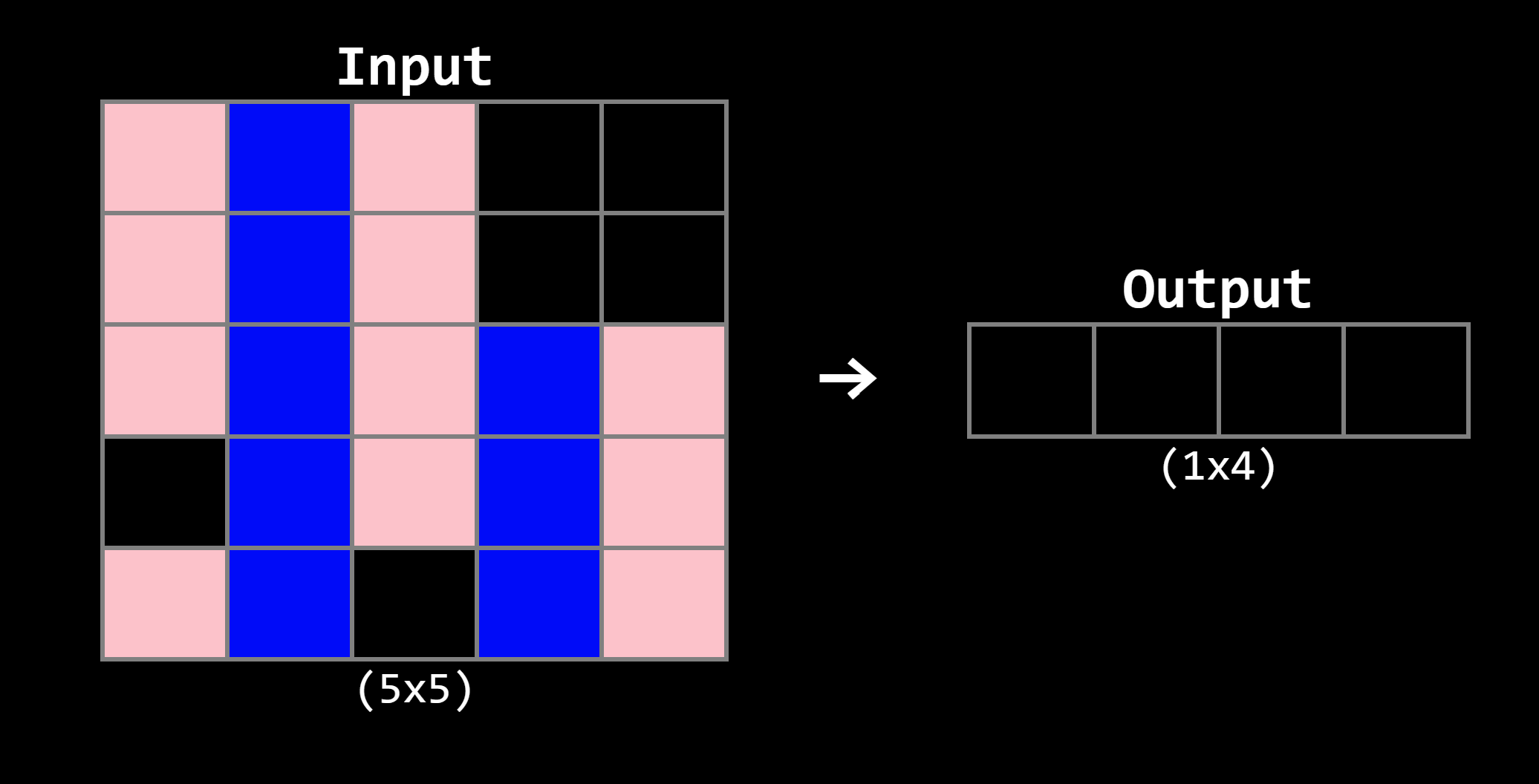}
        \caption{Handling noise from a non-uniform background.}
        \label{fig:hard_count}
    \end{subfigure}
    \caption{Counting Cells: The difficulty increases from simple counting to detecting and analyzing contiguous regions.}
    \label{fig:counting_comparison}
\end{figure*}

\begin{figure*}[t]
    \centering
    \begin{subfigure}{0.32\textwidth}
        \includegraphics[width=\textwidth]{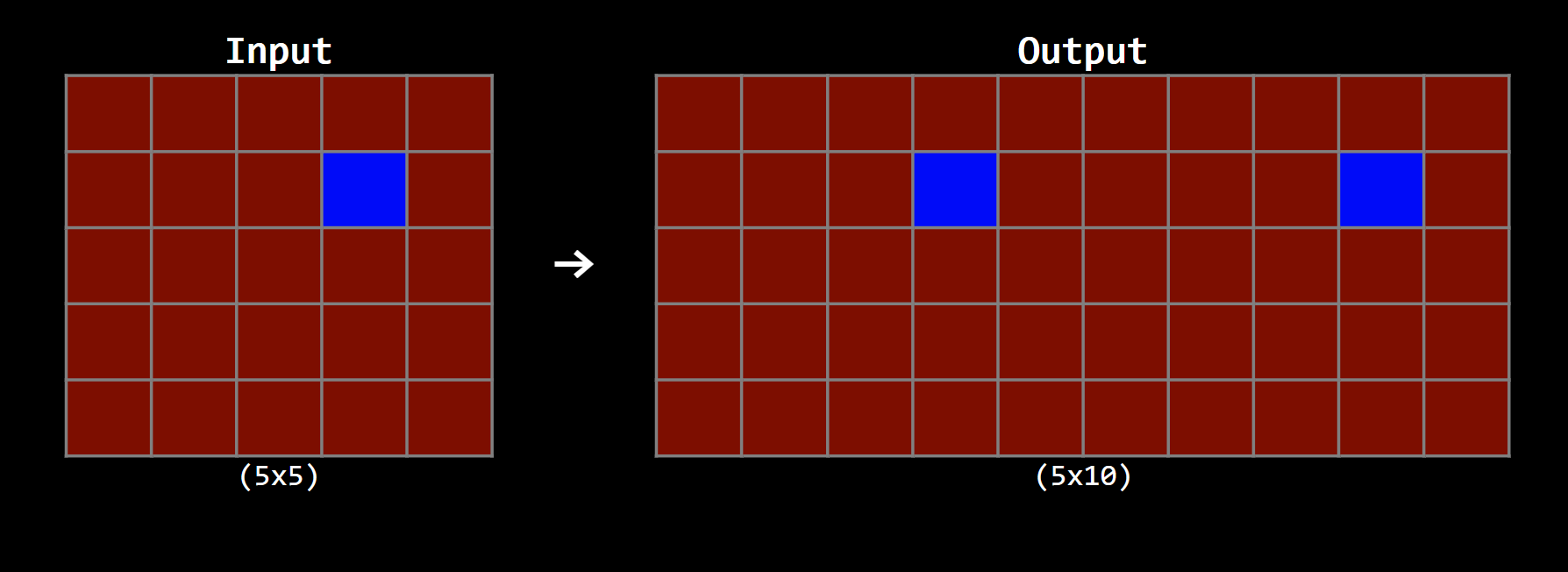}
        \caption{Basic horizontal duplication of a pattern.}
        \label{fig:easy_double}
    \end{subfigure}
    \hfill
    \begin{subfigure}{0.32\textwidth}
        \includegraphics[width=\textwidth]{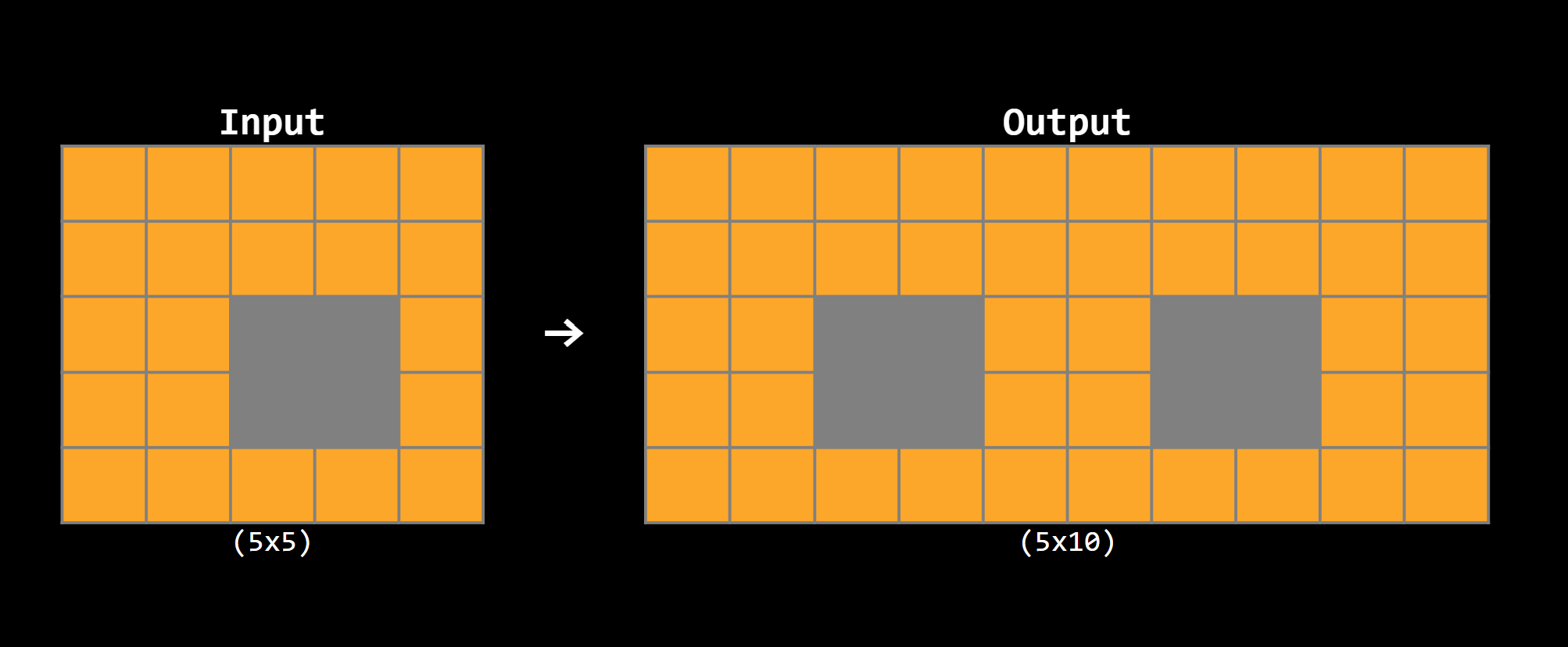}
        \caption{Duplication with vertical flip}
        \label{fig:medium_double}
    \end{subfigure}
    \hfill
    \begin{subfigure}{0.32\textwidth}
        \includegraphics[width=\textwidth]{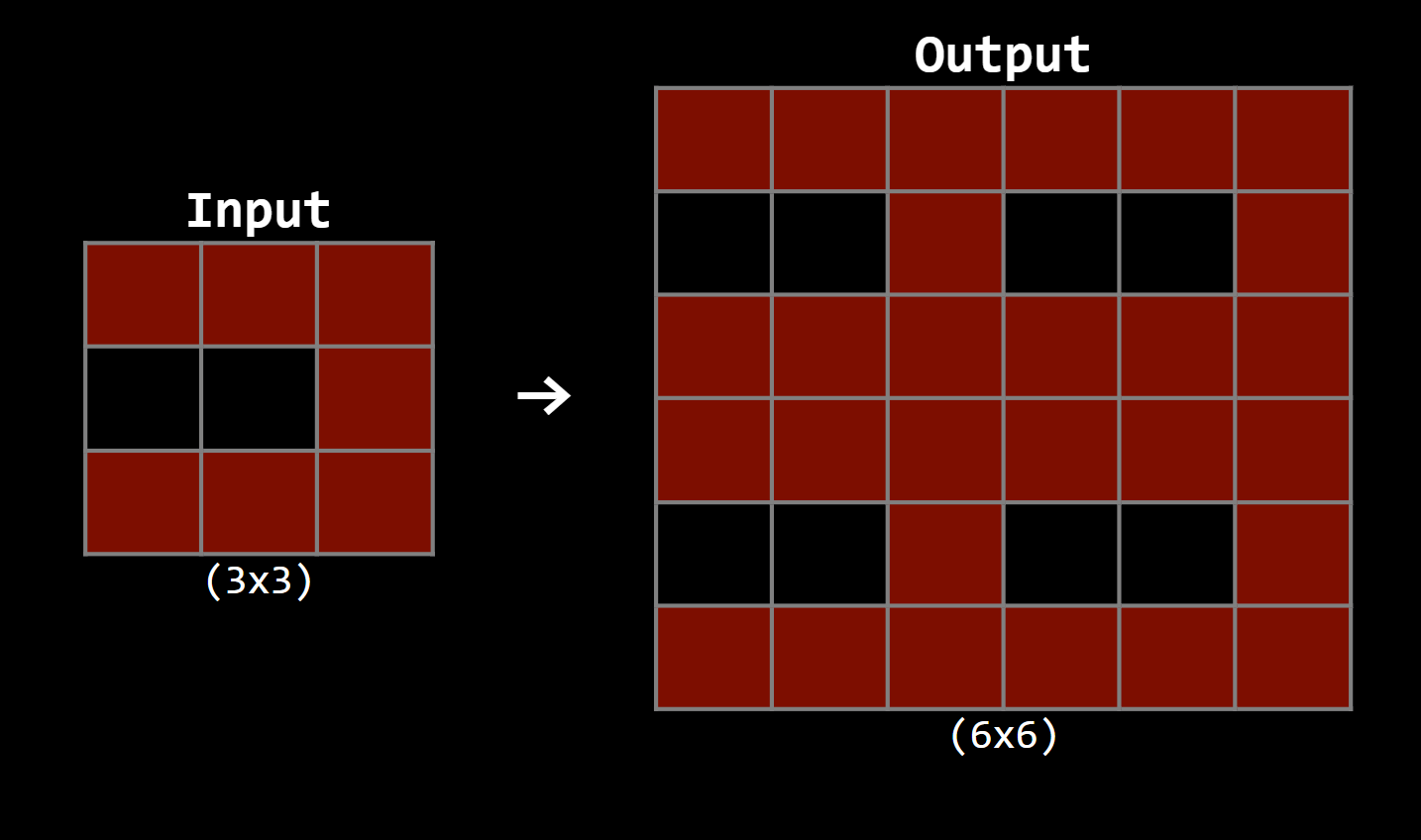}
        \caption{Multiple replications with varied transformations.}
        \label{fig:hard_double}
    \end{subfigure}
    \caption{Double Grid Task: Complexity grows from simple copying to flipping and repeating with transformations.}
    \label{fig:double_comparison}
\end{figure*}

\begin{figure*}[t]
    \centering
    \begin{subfigure}{0.32\textwidth}
        \includegraphics[width=\textwidth]{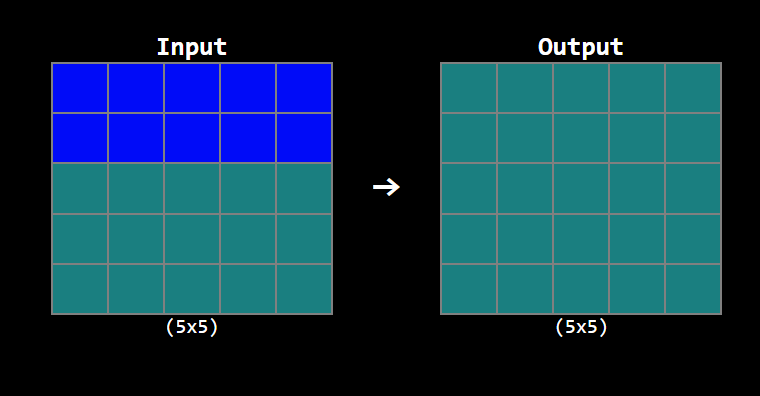}
        \caption{Basic dominant side detection.}
        \label{fig:easy_dominant}
    \end{subfigure}
    \hfill
    \begin{subfigure}{0.32\textwidth}
        \includegraphics[width=\textwidth]{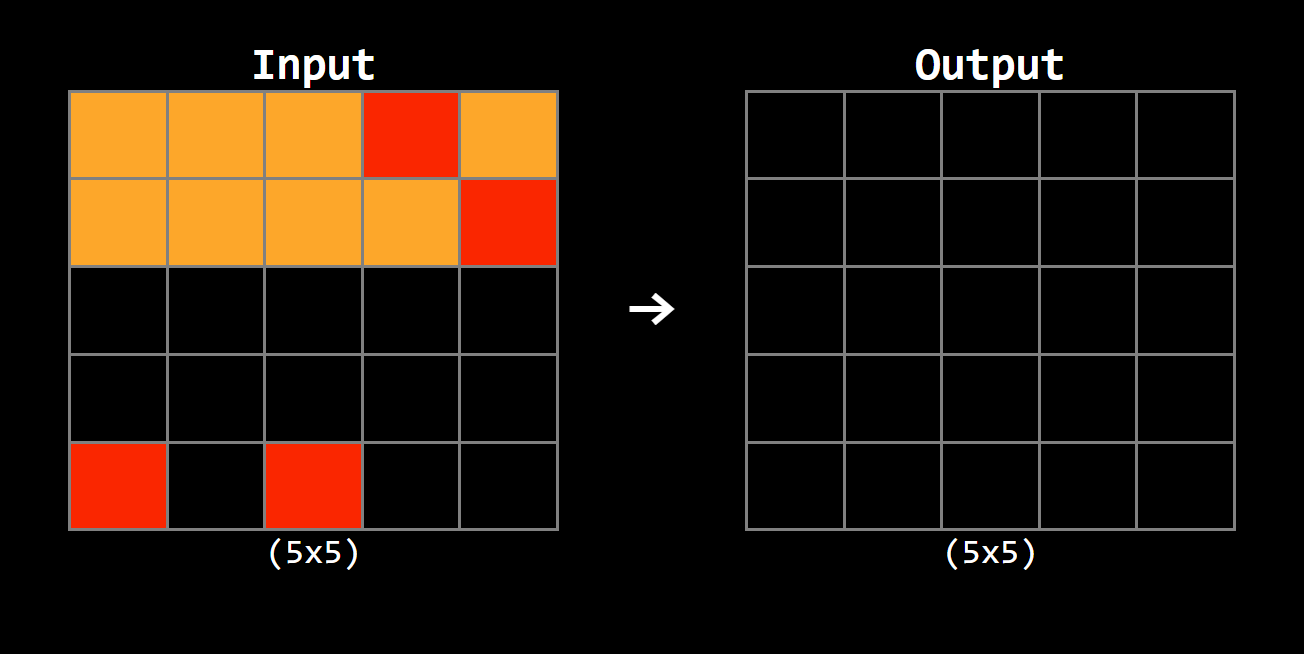}
        \caption{Predicting the dominant region with noise.}
        \label{fig:medium_dominant}
    \end{subfigure}
    \hfill
    \begin{subfigure}{0.32\textwidth}
        \includegraphics[width=\textwidth]{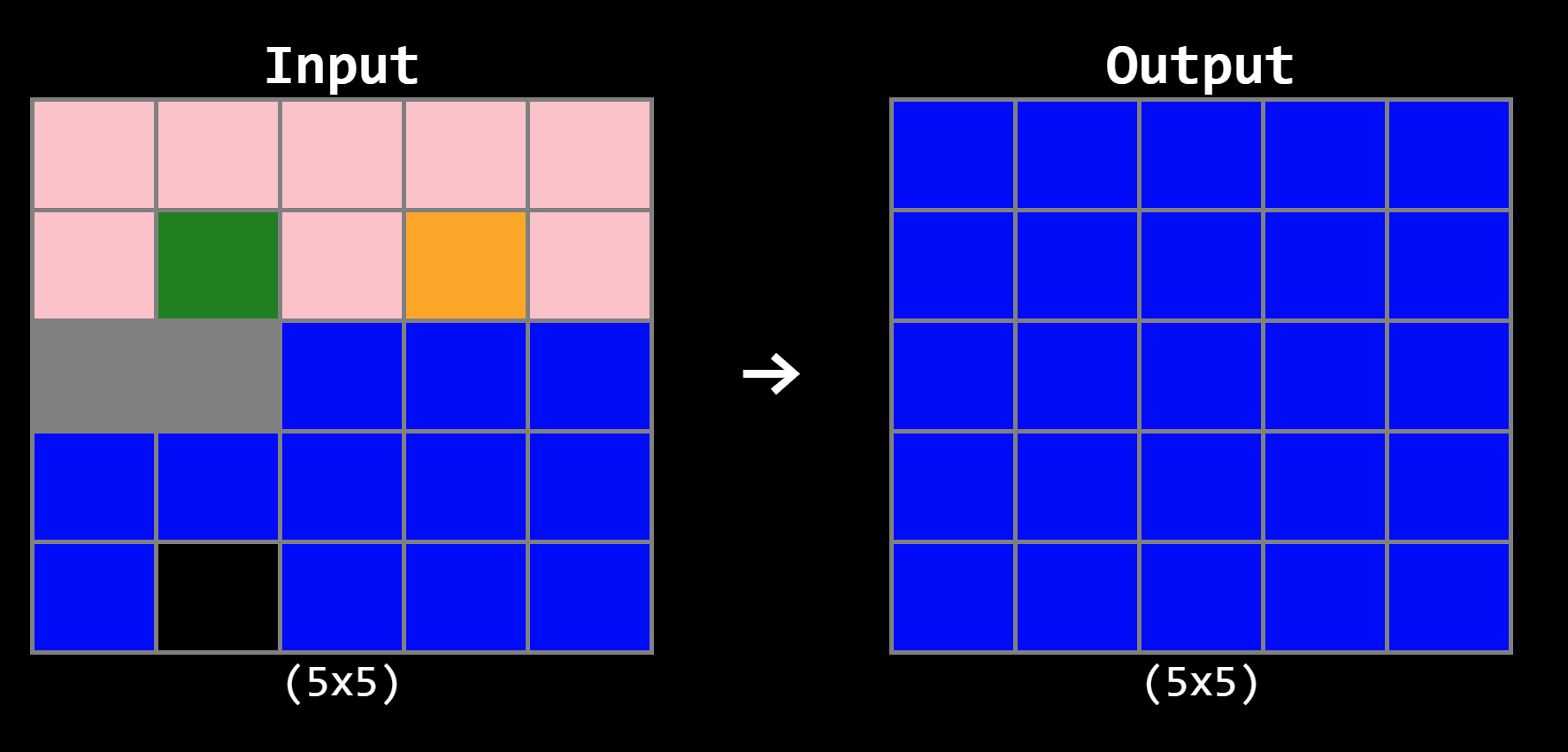}
        \caption{Handling complex noise while detecting dominance.}
        \label{fig:hard_dominant}
    \end{subfigure}
    \caption{Dominant Side Task: Increasing difficulty as noise complicates detection.}
    \label{fig:dominant_comparison}
\end{figure*}
\begin{figure*}[t]
    \centering
    \begin{subfigure}{0.32\textwidth}
        \includegraphics[width=\textwidth]{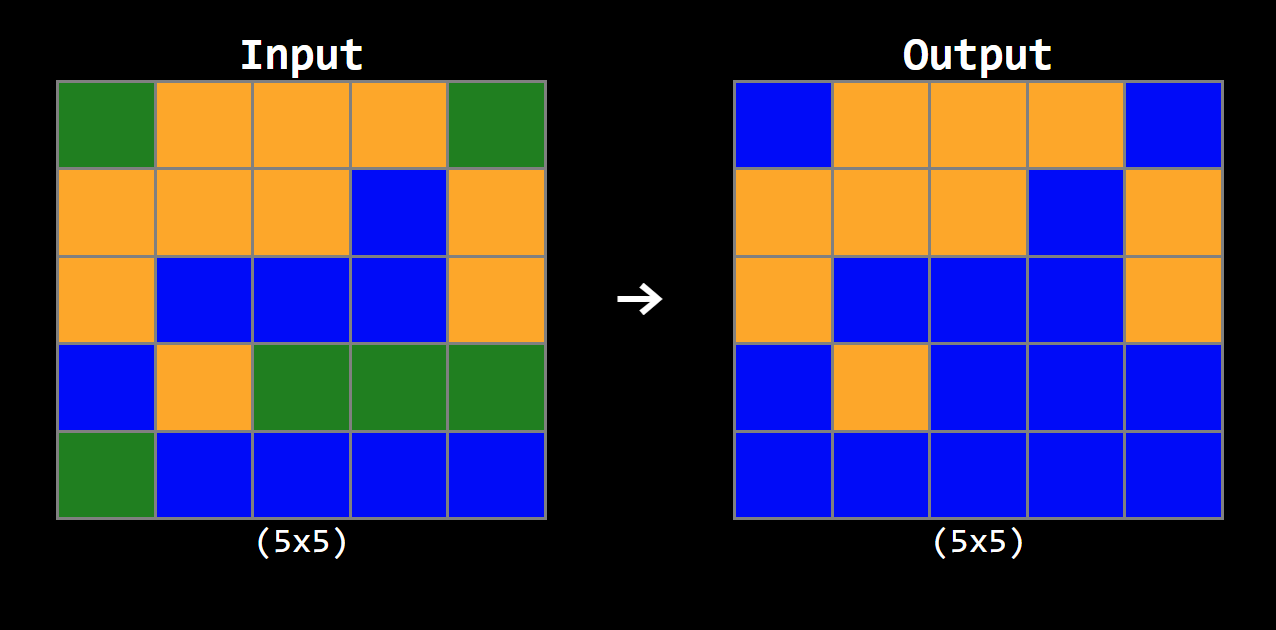}
        \caption{Simple transformation of one color into another across all examples.}
        \label{fig:easy_drop}
    \end{subfigure}
    \hfill
    \begin{subfigure}{0.32\textwidth}
        \includegraphics[width=\textwidth]{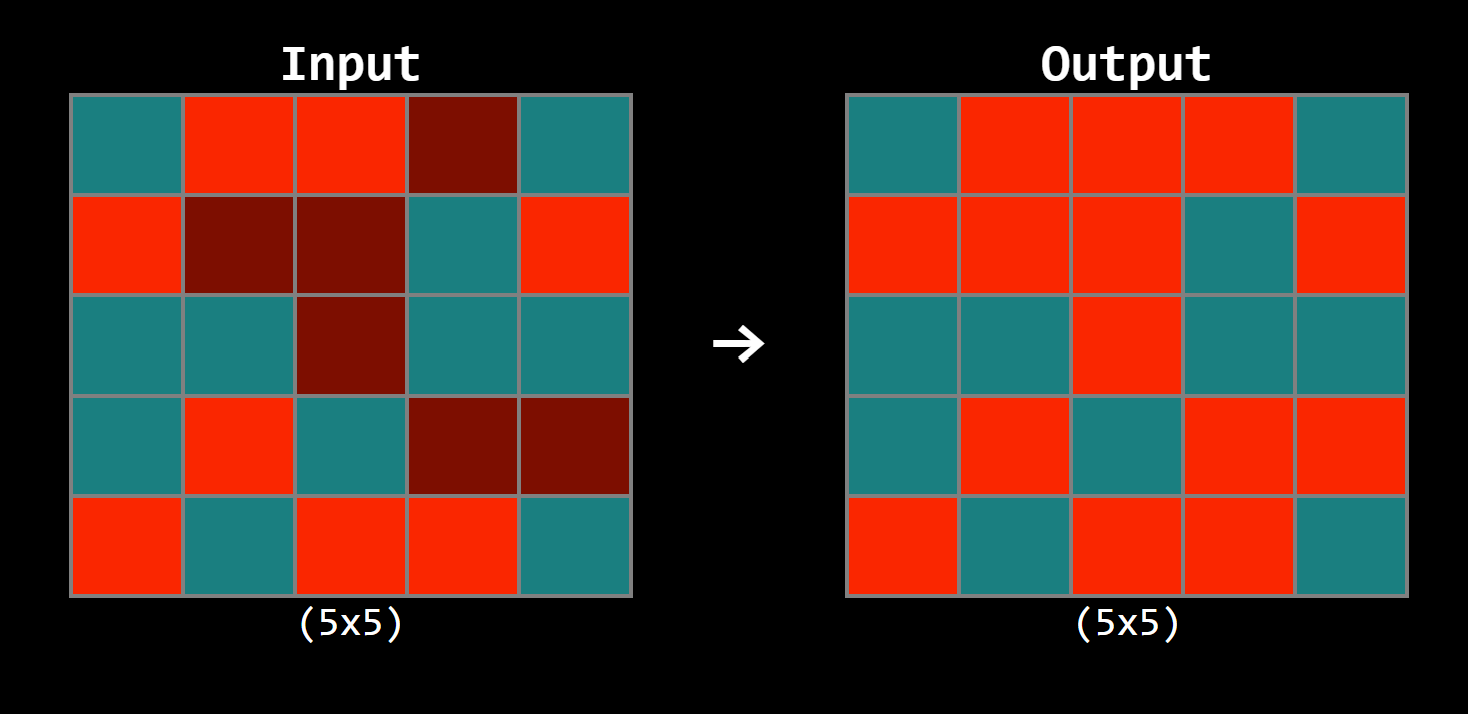}
        \caption{Transformation rule with a varying third color.}
        \label{fig:medium_drop}
    \end{subfigure}
    \hfill
    \begin{subfigure}{0.32\textwidth}
        \includegraphics[width=\textwidth]{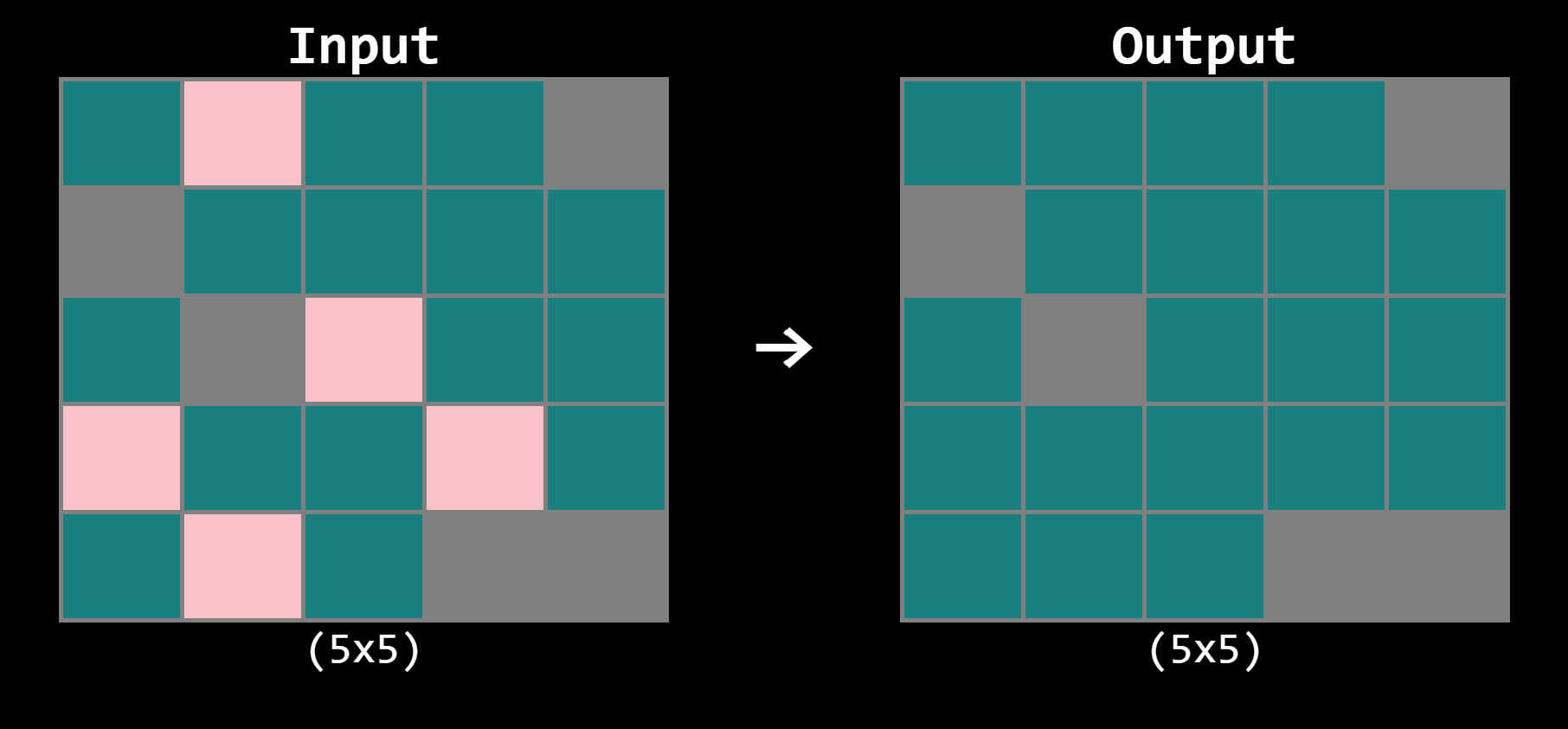}
        \caption{One color blending into a varying background with a varying third color.}
        \label{fig:hard_drop}
    \end{subfigure}
    \caption{Drop One Color Task: Increasing complexity in color transformation and background dynamics.}
    \label{fig:drop_comparison}
\end{figure*}

%% file: main.bbl
\begin{thebibliography}{11}
\providecommand{\natexlab}[1]{#1}
\providecommand{\url}[1]{\texttt{#1}}
\expandafter\ifx\csname urlstyle\endcsname\relax
  \providecommand{\doi}[1]{doi: #1}\else
  \providecommand{\doi}{doi: \begingroup \urlstyle{rm}\Url}\fi

\bibitem[Aky{\"u}rek et~al.(2024)Aky{\"u}rek, Damani, Qiu, Guo, Kim, and Andreas]{akyurek2411surprising}
Ekin Aky{\"u}rek, Mehul Damani, Linlu Qiu, Han Guo, Yoon Kim, and Jacob Andreas.
\newblock The surprising effectiveness of test-time training for abstract reasoning, 2024.
\newblock \emph{Preprint at https://arxiv. org/abs/2411.07279}, 2024.

\bibitem[Antol et~al.(2015)Antol, Agrawal, Lu, Mitchell, Batra, Zitnick, and Parikh]{antol2015vqa}
Stanislaw Antol, Aishwarya Agrawal, Jiasen Lu, Margaret Mitchell, Dhruv Batra, C~Lawrence Zitnick, and Devi Parikh.
\newblock Vqa: Visual question answering.
\newblock In \emph{Proceedings of the IEEE international conference on computer vision}, pages 2425--2433, 2015.

\bibitem[Chen et~al.(2021)Chen, Tang, Qin, Liang, Liu, Xing, and Lin]{chen2021geoqa}
Jiaqi Chen, Jianheng Tang, Jinghui Qin, Xiaodan Liang, Lingbo Liu, Eric~P Xing, and Liang Lin.
\newblock Geoqa: A geometric question answering benchmark towards multimodal numerical reasoning.
\newblock \emph{arXiv preprint arXiv:2105.14517}, 2021.

\bibitem[Chollet(2019)]{chollet2019measure}
Fran{\c{c}}ois Chollet.
\newblock On the measure of intelligence.
\newblock \emph{arXiv preprint arXiv:1911.01547}, 2019.

\bibitem[Cole et~al.(2024)Cole, Osman, Hodel, Duggar, and Scarfe]{cole2024arc}
Jack Cole, Mohamed Osman, Michael Hodel, Keith Duggar, and Tim Scarfe.
\newblock New 50 arc result and current winners interviewed.
\newblock Podcast, 2024.
\newblock Available on Spotify.

\bibitem[Hodel(2024)]{rearc}
Michael Hodel.
\newblock re-arc: Reverse-engineering the abstraction and reasoning corpus.
\newblock \url{https://github.com/michaelhodel/re-arc}, 2024.

\bibitem[Johnson et~al.(2017)Johnson, Hariharan, Van Der~Maaten, Fei-Fei, Lawrence~Zitnick, and Girshick]{johnson2017clevr}
Justin Johnson, Bharath Hariharan, Laurens Van Der~Maaten, Li Fei-Fei, C Lawrence~Zitnick, and Ross Girshick.
\newblock Clevr: A diagnostic dataset for compositional language and elementary visual reasoning.
\newblock In \emph{Proceedings of the IEEE conference on computer vision and pattern recognition}, pages 2901--2910, 2017.

\bibitem[LeGris et~al.(2024)LeGris, Vong, Lake, and Gureckis]{legris2024h}
Solim LeGris, Wai~Keen Vong, Brenden~M Lake, and Todd~M Gureckis.
\newblock H-arc: A robust estimate of human performance on the abstraction and reasoning corpus benchmark.
\newblock \emph{arXiv preprint arXiv:2409.01374}, 2024.

\bibitem[Li and Ellis(2024)]{li2024programming}
Wen-Ding Li and Kevin Ellis.
\newblock Is programming by example solved by llms?
\newblock \emph{arXiv preprint arXiv:2406.08316}, 2024.

\bibitem[Li et~al.(2024)Li, Hu, Larsen, Wu, Alford, Woo, Dunn, Tang, Naim, Nguyen, et~al.]{li2024combining}
Wen-Ding Li, Keya Hu, Carter Larsen, Yuqing Wu, Simon Alford, Caleb Woo, Spencer~M Dunn, Hao Tang, Michelangelo Naim, Dat Nguyen, et~al.
\newblock Combining induction and transduction for abstract reasoning.
\newblock \emph{arXiv preprint arXiv:2411.02272}, 2024.

\bibitem[Lu et~al.(2021)Lu, Gong, Jiang, Qiu, Huang, Liang, and Zhu]{lu2021inter}
Pan Lu, Ran Gong, Shibiao Jiang, Liang Qiu, Siyuan Huang, Xiaodan Liang, and Song-Chun Zhu.
\newblock Inter-gps: Interpretable geometry problem solving with formal language and symbolic reasoning.
\newblock \emph{arXiv preprint arXiv:2105.04165}, 2021.

\end{thebibliography}
